\documentclass{article}

\def \TRtitle{Pareto-Path Multi-Task Multiple Kernel Learning}
\def \TRauthors{Cong Li, Michael Georgiopoulos and Georgios C. Anagnostopoulos}
\def \TRemails{congli@eecs.ucf.edu, michaelg@ucf.edu and georgio@fit.edu}
\def \TRkeywords{Multiple Kernel Learning, Multi-task Learning, Multi-objective Optimization, Pareto Front, Support Vector Machines}

\usepackage{options}
\usepackage{packages}
\usepackage{macros}

\begin{document}

\maketitle

\ifMakeReviewDraft
	\linenumbers
\fi


\begin{abstract}

A traditional and intuitively appealing \ac{MT-MKL} method is to optimize the sum (thus, the average) of objective functions with (partially) shared kernel function, which allows information sharing amongst tasks. We point out that the obtained solution corresponds to a single point on the \ac{PF} of a \ac{MOO} problem, which considers the concurrent optimization of all task objectives involved in the \ac{MTL} problem. Motivated by this last observation and arguing that the former approach is heuristic, we propose a novel \ac{SVM} \ac{MT-MKL} framework, that considers an implicitly-defined set of conic combinations of task objectives. We show that solving our framework produces solutions along a path on the aforementioned \ac{PF} and that it subsumes the optimization of the average of objective functions as a special case. Using algorithms we derived, we demonstrate through a series of experimental results that the framework is capable of achieving better classification performance, when compared to other similar \ac{MTL} approaches.   

\end{abstract}

\section{Introduction}
\label{sec:Introduction}

\acresetall

\ac{MKL} is an important method in kernel learning that has drawn considerable attention since being introduced in \cite{Lanckriet2004}. \ac{MKL} seeks an appropriate kernel function (and, hence, kernel matrix) by linearly or non-linearly combining several pre-selected candidate kernel functions. Given a machine learning problem/task, the optimal kernel matrix is derived by optimizing the associated objective function with respect to the combination coefficients, say $\boldsymbol{\theta} = \left [ \theta_1, \cdots, \theta_M \right ]$, for $M$ pre-specified kernels. Employing \ac{MKL} avoids having to resort to (cross-)validating over different kernel choices, which is computationally cumbersome and may require additional data for validation. A key focus of \ac{MKL} is identifying solutions for $\boldsymbol{\theta}$ by imposing appropriate constraints on it, such as an $L_1$-norm constraint \cite{Lanckriet2004} \cite{Rakotomamonjy2008}, $L_2$-norm constraint \cite{Kloft2008}, and $L_p$-norm constraint with $p>1$ \cite{Kloft2011} as a generalization of the previous two methods. The generalization bound and other theoretical aspects of the $L_p$-norm \ac{MKL} method is extensively studied in \cite{Kloft2012}. Besides searching for the optimal constraints on $\boldsymbol{\theta}$, several other works have been proposed, such as using a Group-Lasso type regularizer \cite{Xu2010} and an $L_1$-norm within-group / $L_s$-norm ($s \geq 1$) group-wise regularizer \cite{Aflalo2011}, nonlinearly combined \ac{MKL} \cite{Cortes2010}, \ac{MKL} with localized $\boldsymbol{\theta}$ \cite{Gonen2008}, \ac{MKL} with hyperkernels \cite{Ong2005}, \ac{MKL} based on the radii of minimum enclosing balls \cite{Gai2010} and other methods, such as the ones of \cite{Rakotomamonjy2008}, \cite{Sonnenburg2006} and \cite{Varma2009}, to name a few. A thorough survey of \ac{MKL} is given in \cite{Gonen2011}.

Another active path of \ac{MKL} research is combining \ac{MKL} with \ac{MTL}, which is commonly referred to as \ac{MT-MKL}. \ac{MTL} aims to simultaneously learn multiple related tasks using shared information, such that each task can benefit from learning all tasks. Existing approaches consider several different types of information sharing strategies. For example, \cite{Aflalo2011}, \cite{Jawanpuria2011} and \cite{Rakotomamonjy2011} applied a mixed-norm regularizer on the weights of each linear model (task), which forces tasks to be related, and, at the same time, achieves different levels of inner-task and inter-task sparsity on the weights. Another example is the model proposed in \cite{Widmer2010}, which considers $T$ tasks and restricts the $T$ \ac{SVM} weights to be close to a common weight, such that the weights from all tasks are related. Additionally, for the recently proposed \textit{Minimax MTL} model \cite{Mehta2012}, tasks are related by minimizing the maximum of the $T$ loss functions, in order to guarantee some minimum level of accuracy for each task. Last but not least, a straightforward strategy of information sharing is to let all tasks share (or partially share) a common kernel function, which has been investigated in \cite{Jebara2004}, \cite{Tang2009}, and \cite{Samek2011}, again to name a few. According to this strategy, tasks are related by mapping data from all tasks to a common feature space and, subsequently, each task is learned in the same, common feature space.

Considering the latter strategy of information sharing, the most intuitively straightforward formulation is to optimize the sum (equivalently, the average) of objective functions with shared kernel function, such as the models in \cite{Tang2009} and \cite{Samek2011}. However, as we subsequently argue, this method is rather an \emph{ad hoc} strategy. First, we observe that optimizing the average of objective functions is equivalent to finding a particular solution to a \ac{MOO} problem, which aims to optimize all task objectives simultaneously. In specific, it is a well known fact (see \cite[p. 178]{Boyd2004}) that scalarizing a \ac{MOO} problem by optimizing different conic combinations (linear combinations with non-negative coefficients) of the objective functions leads to the discovery of solutions that correspond to points on the convex (when minimizing) part of the problem's \ac{PF}. The latter set is the set of non-dominated solutions in the space of objective values. Hence, by optimizing the average of task objectives in an \ac{MTL} setting, one only finds a particular \ac{PF} point  (or more, if the \ac{PF} is non-convex) of the corresponding \ac{MOO} problem. Therefore, while considering the optimization of this average may be intuitively appealing, it is, nevertheless, a largely \emph{ad hoc} strategy. Foreseeably so, optimizing a different conic combination of task objectives, albeit among an infinity of possibilities, may improve the performance of each task even further, when compared to the case of averaged objectives. This amounts to searching for better solutions, \ie\ \ac{PF} points, of the associated \ac{MOO} problem and forms the basis of our work. 

In this paper, we propose a new \ac{SVM}-based \ac{MT-MKL} framework for binary classification tasks. The common kernel utilized by these \ac{SVM} models is established through a typical \ac{MKL} approach. More importantly, though, it considers optimizing specific conic combinations of the task objectives, in order to improve upon the traditional \ac{MTL} method of averaging. In \sref{sec:ProblemFormulation} we show that the obtained solutions trace a path on the \ac{PF} of the relevant \ac{MOO} problem. While doing so does not explore the entire \ac{PF}, searching for solutions to our problem is computationally feasible. The framework's conic combinations and, thus, the aforementioned path, is parameterized by a parameter $p>0$. For $p=1$, the whole framework coincides with the traditional \ac{MTL} approach of minimizing the average of \ac{SVM} objective functions. In \sref{sec:AlgorithmConvex} and \sref{sec:AlgorithmNonconvex} we derive algorithms to solve our proposed framework for $p \geq 1$ and $0 < p < 1$ respectively, while in \sref{sec:Experiments} we demonstrate the impact of $p$'s value on learning performance. In specific, we show that recognition accuracy for all tasks increases as $p$ decreases below $1$. We discuss why this phenomenon occurs and provide insights into the behavior of our \ac{MT-MKL} formulation. In the same section, we also provide a variety of experimental results to highlight the utility of this formulation. Finally, we briefly summarize all our findings in \sref{sec:Conclusions}. 

In the sequel, we'll be using the following notational conventions: vector and matrices are denoted in boldface. Vectors are assumed to be columns vectors. If $\boldsymbol{v}$ is a vector, then $\boldsymbol{v}'$ denotes the transposition of $\boldsymbol{v}$. Vectors $\boldsymbol{0}$ and $\boldsymbol{1}$ are the all-zero and all-one vectors respectively. Also, $\succeq$ and $\max \left\{ \cdot, \cdot \right\}$ between vectors will stand for the component-wise $\geq$ and $\max \left\{ \cdot, \cdot \right\}$ relations respectively. For any $\boldsymbol{v} \succeq \boldsymbol{0}$, $\boldsymbol{v}^p$ represents component-wise exponentiation of $\boldsymbol{v}$. Furthermore, we will be using the notation $\nu( \boldsymbol{v} )_{p} \triangleq ( \boldsymbol{1}' \boldsymbol{v}^p )^\frac{1}{p} $ where $\boldsymbol{v}$ is a vector with $\boldsymbol{v} \succeq \boldsymbol{0}$ and $p \in (0, +\infty]$. Observe that for $p \geq 1$, $\nu( \boldsymbol{v} )_{p} =\left \| \boldsymbol{v} \right \|_p$, where $\left \| \cdot \right \|_p$ stands for the ordinary Minkowski $L_p$-norm for finite-dimensional vectors. Note that for $p \in (0, 1)$, $\nu( \boldsymbol{v} )_{p}$ is not a norm. For any $s > 0$ and vector $\boldsymbol{v}$, we define the set $\bar{B}_{\boldsymbol{v}, s} \triangleq \left \{ \boldsymbol{v}| \boldsymbol{v} \succeq \boldsymbol{0},  \nu \left ( \boldsymbol{v} \right )_s \leq 1 \right \}$. Also, let $Z_n$ be the set of integers $\left\{1,\cdots,n \right\}$ for any $n \geq 1$. Finally, notice that the proofs of the manuscript's theoretical results are provided in the Appendix.

\section{Problem Formulation}
\label{sec:ProblemFormulation}

Consider the following \ac{MT-MKL} problem involving $T$ \ac{SVM} training tasks:

\begin{equation}
	\begin{aligned}
	\min_{\boldsymbol{f}, \boldsymbol{\theta}, \boldsymbol{\xi}, \boldsymbol{b}} & \nu ( \boldsymbol{g} (\boldsymbol{f}, \boldsymbol{\theta}, \boldsymbol{\xi}) )_p \\
	\textit{s.t.} \;\; & y_i^t(\sum_{m=1}^M f_m^t(\boldsymbol{x}_i^t) + b^t) \geq 1 - \xi_i^t, \;\; \forall i \in Z_{N_t}, t \in Z_T \\
	& \boldsymbol{\xi}^t \succeq \boldsymbol{0}, \;\; \forall t \in Z_T \\
	& \boldsymbol{\theta} \in \bar{B}_{\boldsymbol{\theta}, s}, \;\; s \geq 1 
	\end{aligned}
	\label{pr:PFOne}
\end{equation}

\noindent 
where $\boldsymbol{g} ( \boldsymbol{f}, \boldsymbol{\theta}, \boldsymbol{\xi} ) \triangleq [ g^1(\boldsymbol{f}, \boldsymbol{\theta}, \boldsymbol{\xi}), \cdots, g^T(\boldsymbol{f}, \boldsymbol{\theta}, \boldsymbol{\xi})]'$ and each $g^t(\boldsymbol{f}, \boldsymbol{\theta}, \boldsymbol{\xi})$ is defined as the $t$-th multi-kernel \ac{SVM} objective, \ie.

\begin{equation}
g^t(\boldsymbol{f}, \boldsymbol{\theta}, \boldsymbol{\xi}) \triangleq \sum_{m=1}^M\frac{\left \| f_m^t \right \|_{H_m}^2}{2\theta_m} + C \sum_{i=1}^{N_t}\xi_i^t
\label{pr:GFunction}
\end{equation}

\noindent
where $\boldsymbol{f} \triangleq [ {\boldsymbol{f}^1}',\cdots,{\boldsymbol{f}^T}' ]'$, $\boldsymbol{f}^t \triangleq [ f_1^t, \cdots, f_M^t ]'$,  $\boldsymbol{\xi} \triangleq [ {\boldsymbol{\xi}^1}',\cdots,{\boldsymbol{\xi}^T}' ]'$, $\boldsymbol{\xi}^t \triangleq [ \xi^t_1,\cdots, \xi^t_{N^t} ]'$, $\boldsymbol{\theta} \triangleq [ \theta_1,\cdots,\theta_M ]'$, $\boldsymbol{b} \triangleq [ b^1, \cdots, b^T ]'$. Moreover, $\left \{ \boldsymbol{x}_i^t, y_i^t \right \}$, where $i=1,\cdots , N^t$, are the training samples available for the $t^{th}$ task.

For each task $t$, $M$ discriminative functions $f_m^t$ are sought under the constraints $f_m^t \in \mathcal{H}_m$, where each $\mathcal{H}_m$ is a \ac{RKHS} associated to a pre-selected reproducing kernel $k_m(\cdot, \cdot)$. Let $\boldsymbol{x}$ encompass all the optimization variables, that is, $\boldsymbol{f}, \boldsymbol{\theta}, \boldsymbol{\xi}, \boldsymbol{b}$. We can restate \pref{pr:PFOne} as

\begin{equation}
	\min_{\boldsymbol{x} \in \Omega ( \boldsymbol{x} )} \nu ( \boldsymbol{g} ( \boldsymbol{x} ) )_p 
	\label{pr:PFTwo}
\end{equation}

\noindent 
where $\Omega ( \boldsymbol{x} )$ is the feasible region of $\boldsymbol{x}$, given by the constraints of \pref{pr:PFOne}. Note that the \ac{SVM} objective functions $\boldsymbol{g} ( \boldsymbol{x} )$ are non-negative and not all simultaneously zero for any value of $\boldsymbol{x}$. Therefore, $\nu ( \boldsymbol{g} ( \boldsymbol{x} ) )_p$ is well defined, based on the definition of $v(\cdot)_p$ given in \sref{sec:Introduction}.

Since $\nu(\cdot)_1 = \left\| \cdot \right\|_1$, the traditional \ac{MT-MKL} formulation considers only the case, when $p = 1$, \ie, the sum (or, equivalently, the average) of the $T$ objectives. However, as argued in \sref{sec:Introduction}, optimizing the objectives' average may in practice not necessarily lead to achieving the best obtainable performance for every task simultaneously and, thus, it is of interest to investigate cases, for which $p \neq 1$. We show that, for any $p > 0$, the optimum value $\boldsymbol{g}^*$ is a \ac{PF} solution for the $T$ \ac{SVM} objectives. Based on this conclusion, we are able to explore a path on the \ac{PF} by tuning only one parameter, namely $p$. We later discuss that by doing so, it not only helps us achieve uniform performance improvements, but it also provides useful insights into the \ac{SVM}-based {MT-MKL} problem we are considering.

\begin{proposition}
\label{prop:ParetoPoints}

For any $p > 0$ and arbitrary vector function $\boldsymbol{g}(\boldsymbol{x}) \in \mathbb{R}^T$ with $\boldsymbol{g} \succeq \boldsymbol{0}$ for all vectors $\boldsymbol{x}$ in a feasible set $\Omega ( \boldsymbol{x} )$, the optimal solution $\boldsymbol{x}^*$ of the general optimization problem

\begin{equation}
	\min_{\boldsymbol{x}\in \Omega ( \boldsymbol{x} )}  \nu ( \boldsymbol{g} ( \boldsymbol{x} ) )_p 
	\label{pr:PFThree}
\end{equation}

\noindent 
is a \ac{PF} solution of the \ac{MOO} problem

\begin{equation}
	\min_{\boldsymbol{x}\in \Omega ( \boldsymbol{x} )}  \boldsymbol{g} ( \boldsymbol{x} )  
	\label{pr:MOO}
\end{equation}

\noindent which considers the simultaneous minimization of the $T$ objectives $g_1 ( \boldsymbol{x} ),\cdots,g_T ( \boldsymbol{x} )$.
\end{proposition}

\noindent 
The proof is given in \sref{sec:ProofParetoPoints} of the Appendix. It is readily evident that our framework is convex, when $p \geq 1$, and non-convex, when $p < 1$. In the following two sections, we discuss how to optimize \pref{pr:PFOne} in both cases. 

\section{Learning in the Convex Case}
\label{sec:AlgorithmConvex}

For $p \geq 1$, we first convert \pref{pr:PFOne} to an equivalent min-max optimization problem and then employ a specialized version of an algorithm proposed in \cite{Tseng2008} to solve it.

\begin{lemma}
\label{lemm:EquivProblemConvexCase}

Let $p \geq 1$, $\boldsymbol{\lambda}, \boldsymbol{g} \in \mathbb{R}^T$ such that $\boldsymbol{g} \succeq \boldsymbol{0}$, but $\boldsymbol{g} \neq \boldsymbol{0}$. Also, let $q \triangleq \frac{p}{p-1}$. Then,

\begin{equation}
	\max_{\boldsymbol{\lambda} \in \bar{B}_{\boldsymbol{\lambda}, q}} \boldsymbol{\lambda}' \boldsymbol{g} = \nu(\boldsymbol{g})_p = \left \| \boldsymbol{g} \right \|_p
	\label{eq:ACOne}
\end{equation}

\noindent
Furthermore, a solution to the previously stated maximization problem is given as

\begin{equation}
	\boldsymbol{\lambda}^* = 
		\begin{cases} 
				( \frac{\boldsymbol{g}}{ \left \| \boldsymbol{g} \right \|_p } )^{p-1}  & \text{ if } p>1 \\ 
				\boldsymbol{1} & \text{ if } p=1 \\		
		\end{cases} 
	\label{eq:ACTwo}
\end{equation} 
\end{lemma}

\noindent
The veracity of the previous lemma is easily demonstrated. In specific, note that \eref{eq:ACOne} is similar to the definition of the dual norm (see \cite[p. 637]{Boyd2004}):

\begin{equation}
	\left \| \boldsymbol{g} \right \|_* = \sup_{\left \| \boldsymbol{\lambda} \right \| \leq 1 }\boldsymbol{\lambda}' \boldsymbol{g}
	\label{eq:dual_norm}
\end{equation}

\noindent
where $\left \| \cdot  \right \|_*$ is the dual norm of $\left \| \cdot  \right \|$. A slight difference between \pref{eq:ACOne} and \pref{eq:dual_norm} is that, in \pref{eq:ACOne}, the constraint is $\boldsymbol{\lambda} \in \bar{B}_{\boldsymbol{\lambda}, q}$ (\ie\ $\boldsymbol{\lambda} \succeq \boldsymbol{0}$ and $\left \| \boldsymbol{\lambda} \right \|_q \leq 1$). However, note that, as $\boldsymbol{g} \succeq \boldsymbol{0}$, the optimal $\boldsymbol{\lambda}$ must satisfy $\boldsymbol{\lambda} \succeq \boldsymbol{0}$. Therefore, under the condition $\boldsymbol{g} \succeq \boldsymbol{0}$, (\ref{eq:ACOne}) is the same as the definition of the dual norm. Finally, \eref{eq:ACTwo} gives the solution of the maximization problem in \eref{eq:ACOne} and its correctness can be verified by directly substituting \eref{eq:ACTwo} into \pref{eq:ACOne}. 

The previous lemma implies that, for $p \geq 1$, instead of optimizing $\nu(\boldsymbol{g})_p$, one could optimize the conic combination of objective functions $g_t$ with coefficients $\boldsymbol{\lambda}^*$ as given in \eref{eq:ACTwo}. Furthermore, it allows us to transform \pref{pr:PFOne} into an equivalent, more suitable problem via the next theorem.

\begin{theorem}
\label{thm:EquivProblemConvexCase}

For $p \geq 1$, \ac{MT-MKL} optimization \pref{pr:PFOne} is equivalent to the following min-max problem

\begin{equation}
\begin{aligned}
 & \min_{\boldsymbol{\theta}} \max_{\boldsymbol{\lambda}, \boldsymbol{\beta}}  \;\; \Phi (\boldsymbol{\theta}, \boldsymbol{\beta}, \boldsymbol{\lambda} )\\
& \text{s.t.} \; \; {\boldsymbol{\beta}^{t}}'\boldsymbol{y}^t=0, \; \; \lambda^t C \boldsymbol{1} \succeq \boldsymbol{\beta}^{t} \succeq \boldsymbol{0}, \; \; t \in Z_T; \\
& \; \; \; \; \; \boldsymbol{\theta} \in \bar{B}_{\boldsymbol{\theta}, s}, \;\; s \geq 1; \; \;  \boldsymbol{\lambda} \in \bar{B}_{\boldsymbol{\lambda}, q}.
\end{aligned}
\label{pr:ACNine}
\end{equation}

\noindent
where $\Phi (\boldsymbol{\theta}, \boldsymbol{\beta}, \boldsymbol{\lambda} ) \triangleq \sum_{t=1}^T ( {\boldsymbol{\beta}^{t}}' \boldsymbol{1} - \frac{1}{2 \lambda^t}{\boldsymbol{\beta}^{t}}'{\boldsymbol{Y}^{t}}'  $ $ (  \sum_{m=1}^M  \theta_m \boldsymbol{K}_m^t  )  \boldsymbol{Y}^t \boldsymbol{\beta}^{t} ) $ and $q \triangleq \frac{p}{p-1}$.
\end{theorem}

\noindent
The proof of the above theorem is given in \sref{sec:ProofThmEquivProblemConvexCase} of the Appendix. In \pref{pr:ACNine}, $\beta_i^t \triangleq \alpha_i^t \lambda^t$, where $\alpha_i^t$'s are the dual variables of the $t$-th \ac{SVM} problem, $\boldsymbol{y}^t \in \left\{ -1, 1\right\}^{N_t}$ is the vector containing all labels of the training data for the $t$-th task, $\boldsymbol{Y}^t$ is a diagonal matrix with the elements of $\boldsymbol{y}^t$ on its diagonal, and $\boldsymbol{K}_m^t$ is the kernel matrix with elements $k_m ( \boldsymbol{x}_i^t,\boldsymbol{x}_j^t )$.

\subsection{Tseng's Algorithm}
\label{sec:AlgorithmConvexSubSec}

Note that \pref{pr:ACNine} is a convex-concave optimization problem. Based on this fact, we consider Tseng's algorithm \cite{Tseng2008} for solving the problem. Define $\boldsymbol{u} \triangleq { [ {\boldsymbol{\theta}}', {\boldsymbol{\beta}}', \boldsymbol{\lambda}' ]}'$ and let $\Phi ( \boldsymbol{u} ) \triangleq \Phi ( \boldsymbol{\theta}, \boldsymbol{\beta}, \boldsymbol{\lambda} )$ stand for the objective function of \pref{pr:ACNine}. Moreover, the algorithm considers the vector function $\boldsymbol{q}  ( \boldsymbol{u}  ) \triangleq [ \nabla_{\boldsymbol{\theta}}\Phi ( \boldsymbol{u} )', -\nabla_{\boldsymbol{\beta}}\Phi ( \boldsymbol{u}  )', -\nabla_{\boldsymbol{\lambda}}\Phi ( \boldsymbol{u} )' ]'$. During the $k$-th iteration, assuming that $\boldsymbol{u}_k$ is already known, the algorithm finds $\zeta > 0$, such that $\boldsymbol{v}_k$, which is given by 

\begin{equation}
\boldsymbol{v}_k =  \arg \min_{\boldsymbol{u} \in \Omega \left( \boldsymbol{u} \right)} \left \{ \boldsymbol{u}' \boldsymbol{q} ( \boldsymbol{u}_k ) + \zeta D\left ( \boldsymbol{u}, \boldsymbol{u}_k \right ) \right \} 
\label{eq:ACTwelve} 
\end{equation}

\noindent 
satisfies the following condition:

\begin{equation}
\min_{\boldsymbol{u} \in \Omega \left( \boldsymbol{u} \right)} \left \{  \boldsymbol{u}' \boldsymbol{q} (\boldsymbol{v}_k) + \zeta D\left ( \boldsymbol{u}, \boldsymbol{u}_k \right ) \right \} \geq \boldsymbol{v}'_k \boldsymbol{q} (\boldsymbol{v}_k )  
\label{pr:ACEleven} 
\end{equation}

\noindent  
Subsequently, $\boldsymbol{u}_{k+1}$ is set as the minimizer of problem (\ref{pr:ACEleven}). In the last two problems, $D(\cdot,\cdot)$ denotes the Bregman divergence, which, for any strictly convex function $h$, is defined as $D ( \boldsymbol{u}, \boldsymbol{v} ) = h ( \boldsymbol{u} ) - h ( \boldsymbol{v} ) -  ( \boldsymbol{u}-\boldsymbol{v} )' \nabla h ( \boldsymbol{v}  )$.  $\Omega ( \boldsymbol{u} )$ is the feasible region of $\boldsymbol{u}$; for our problem it is the feasible set of $\boldsymbol{\theta}$, $\boldsymbol{\beta}$ and $\boldsymbol{\lambda}$ given by the constraints in \pref{pr:ACNine}.  To find $\zeta$, the authors in \cite{Nemirovski2005} suggest initializing $\zeta$ to a large positive value and then to keep halving it until (\ref{pr:ACEleven}) is satisfied. Tseng's algorithm successfully converges, when $\Phi$ is convex-concave and differentiable with Lipschitz-continuous gradient $\boldsymbol{q}$. This condition is not satisfied, when $\exists \ t$ such that $\lambda_t = 0$. However, we will show that this will never happen to our algorithm and, thus, it does not affect its convergence.

\subsection{Adaptation of Tseng's Algorithm to our Framework}
\label{sec:implementationdetails}

In order to solve \pref{pr:ACNine} with Tseng's algorithm, which consists of solving \pref{eq:ACTwelve} and the minimization problem on the left side of \pref{pr:ACEleven}, we first show how to solve \pref{eq:ACTwelve} in our setting. We choose $h( \boldsymbol{u}) = h_{\boldsymbol{\theta}}( \boldsymbol{\theta}_{\boldsymbol{u}}) + h_{\boldsymbol{\beta}}( \boldsymbol{\beta}_{\boldsymbol{u}}) + h_{\boldsymbol{\lambda}}( \boldsymbol{\lambda}_{\boldsymbol{u}})$, where $h_{\boldsymbol{\beta}} ( \boldsymbol{\beta}  ) \triangleq  \| \boldsymbol{\beta}  \|_2^2$, $h_{\boldsymbol{\lambda}} ( \boldsymbol{\lambda}  ) \triangleq  \| \boldsymbol{\lambda}  \|_2^2$, and $h_{\boldsymbol{\theta}} ( \boldsymbol{\theta} ) \triangleq \frac{1}{\bar{s}} \| \boldsymbol{\theta}  \|_{\bar{s}}^{\bar{s}}$ with $\bar{s} = s$, when $s > 1$, and $\bar{s} = 2$, when $s = 1$. Given our choices, it is not difficult to see that the minimizations in \pref{eq:ACTwelve} can be separated into the following two problems: 

\begin{equation}
	 \min_{\boldsymbol{\lambda}, \boldsymbol{\beta} \in \Omega (\boldsymbol{\lambda}, \boldsymbol{\beta} )}  \left \| \boldsymbol{\beta} \right \|_2^2 - {\boldsymbol{\beta}}'  ( \frac{1}{\zeta}\nabla_{\boldsymbol{\beta}}\Phi ( \boldsymbol{u}_k ) + 2\boldsymbol{\beta}_{\boldsymbol{u}_k} ) + \left \| \boldsymbol{\lambda} \right \|_2^2 - {\boldsymbol{\lambda}}' ( \frac{1}{\zeta}\nabla_{\boldsymbol{\lambda}}\Phi ( \boldsymbol{u}_k ) +2\boldsymbol{\lambda}_{\boldsymbol{u}_k} )
	\label{pr:ACTwentyTwo} 
\end{equation}

\begin{equation}
	\min_{\boldsymbol{\theta} \in \Omega ( \boldsymbol{\theta} )} \;\; \frac{1}{\bar{s}} \left \| \boldsymbol{\theta} \right \|_{\bar{s}}^{\bar{s}} - {\boldsymbol{\theta}}'  ( \boldsymbol{\theta}_{\boldsymbol{u}_k}^{\bar{s}-1} - \frac{1}{\zeta} \nabla_{\boldsymbol{\theta}}\Phi ( \boldsymbol{u}_k  ) )
	\label{pr:ACTwentyThree}
\end{equation}

\noindent 
\pref{pr:ACTwentyTwo} is a convex optimization problem, which can be solved via many general-purpose optimization tools, such as \texttt{cvx} \cite{cvx}\cite{Grant2008}. On the other hand, \pref{pr:ACTwentyThree} has a closed-form solution that is provided by the following theorem.

\begin{theorem}
\label{thm:SpecialMinProb1}
Let $\boldsymbol{\theta} \in \mathbb{R}^M$ and $\boldsymbol{\psi} \in \mathbb{R}^M$ such that $\boldsymbol{\psi} \succeq \boldsymbol{0}$, $s > 1$ and $r \triangleq \frac{1}{s-1}$. The unique solution of the constrained minimization problem

\begin{equation}
	\min_{\boldsymbol{\theta} \in \Omega \left( \boldsymbol{\theta} \right)} \frac{1}{s} \boldsymbol{1}' \boldsymbol{\theta}^s -  \boldsymbol{\psi}' \boldsymbol{\theta}
	\label{pr:ACTwentyFour}
\end{equation} 

\noindent
is given as

\begin{equation}
\boldsymbol{\theta}^* = \frac{\boldsymbol{\psi}^r}{\max \left \{ 1, \left \| \boldsymbol{\psi}^r \right \|_s \right \}}
\label{eq:ACTwentyFive}
\end{equation}

\noindent if $\Omega \left( \boldsymbol{\theta} \right) \triangleq \left \{ \boldsymbol{\theta} | \boldsymbol{\theta} \in \bar{B}_{\boldsymbol{\theta}, s} \right \}$ and

\begin{equation}
\boldsymbol{\theta}^* = (\max \left\{\boldsymbol{\psi}-\mu \boldsymbol{1}, \boldsymbol{0} \right\} )^r
\label{eq:ACTwentySix}
\end{equation} 

\noindent if $\Omega \left( \boldsymbol{\theta} \right) \triangleq \left \{ \boldsymbol{\theta} | \boldsymbol{\theta} \in \bar{B}_{\boldsymbol{\theta}, 1}  \right \}$. In \eref{eq:ACTwentySix}, $\mu$ is the smallest nonnegative real number, such that $\left \|\boldsymbol{\theta}^* \right \|_1 = 1$.

\end{theorem}

\noindent 
Solving the minimization problem in (\ref{pr:ACEleven}) can be accomplished by utilizing the similar procedure as the one for solving \pref{eq:ACTwelve}, since these two problems have the same form. The algorithm is deemed to have converged, when the duality gap 
$\max_{\boldsymbol{\beta}, \boldsymbol{\lambda} \in \Omega (\boldsymbol{\beta}, \boldsymbol{\lambda} )} \Phi ( \boldsymbol{\theta}_{\boldsymbol{u}_k}, \boldsymbol{\beta}, \boldsymbol{\lambda} ) - \min_{\boldsymbol{\theta} \in \Omega ( \boldsymbol{\theta} )} \Phi ( \boldsymbol{\theta}, \boldsymbol{\beta}_{\boldsymbol{u}_k}, \boldsymbol{\lambda}_{\boldsymbol{u}_k} )$ is smaller than a predefined threshold. The algorithm is summarized in \aref{alg:Algorithm1}.

\begin{algorithm}
\caption{Algorithm when $p \geq 1$}
\label{alg:Algorithm1}
\begin{algorithmic}
\STATE Choose $M$ kernel functions. Calculate the kernel matrices $\boldsymbol{K}_m^t$ for the $T$ tasks and the $M$ kernels. Initialize $\zeta$, $\boldsymbol{u}_0 \in \Omega \left(\boldsymbol{u} \right)$, $\epsilon$, and $k=0$. 
\WHILE{The duality gap is larger than $\epsilon$}
\STATE Given $\boldsymbol{u}_k$, solve \pref{eq:ACTwelve} and get $\boldsymbol{v}_k$;
\STATE Solve the minimization problem in (\ref{pr:ACEleven});
\IF {The inequality (\ref{pr:ACEleven}) is not satisfied}
\STATE $\zeta \gets \frac{\zeta}{2}$;
\ELSE
\STATE Set $\boldsymbol{u}_{k+1}$ as the optimal of (\ref{pr:ACEleven}); 
\STATE $k \gets k+1$; 
\ENDIF
\ENDWHILE

\end{algorithmic}
\end{algorithm}

As discussed before, in order for the algorithm to converge, we must have $\boldsymbol{\lambda} \succ \boldsymbol{0}$ in every iteration. We show that this is always the case. First, notice that we solve \pref{pr:ACTwentyTwo} to update $\boldsymbol{\lambda}$. Since $\left ( \frac{1}{\zeta}\bigtriangledown_{\boldsymbol{\lambda}}\Phi\left ( \boldsymbol{v}_k \right ) +2\boldsymbol{\lambda}_{\boldsymbol{u}_k} \right ) \succ \boldsymbol{0}$, if $\boldsymbol{\lambda}_{\boldsymbol{v}_k} \succ \boldsymbol{0}$, the optimum solution satisfies $\boldsymbol{\lambda}_{k+1} \neq \boldsymbol{0}$. Therefore, if we initialize $\boldsymbol{\lambda}$, such that $\boldsymbol{\lambda} \succ \boldsymbol{0}$, it will hold that $\boldsymbol{\lambda}_k \succ \boldsymbol{0}$ for all iterations. Secondly, it is not difficult to see from \lemmaref{lemm:EquivProblemConvexCase} that the optimal solution satisfies $\boldsymbol{\lambda}^* \succ \boldsymbol{0}$ for $p < \infty$. Therefore, $\boldsymbol{\lambda}_k$ will safely converge to the optimum.

Note that, besides the algorithm we just introduced, when $p = 1$ we can optimize the model using block-coordinate descent as an alternative. In this case, the framework reduces to the traditional \ac{MT-MKL} approach:

\begin{equation}
\begin{aligned}
\min_{\boldsymbol{x} \in \Omega\left ( \boldsymbol{x} \right )}  \;\; \boldsymbol{g}'\left( \boldsymbol{x}\right)  \boldsymbol{1}
\label{pr:non_convex_opt_traditional_MTMKL}
\end{aligned}
\end{equation}

\noindent
We can optimize with respect to $\left \{ \boldsymbol{f}, \boldsymbol{\xi}, \boldsymbol{b} \right \}$ as a group, which involves $T$ \ac{SVM} problems, and then with respect to $\boldsymbol{\theta}$. The later parameter can be solved for via a closed-form expression.

\section{Learning in the Non-Convex Case}
\label{sec:AlgorithmNonconvex}

In this section we provide a simple algorithm to solve our framework in the case, when $p \in ( 0, 1 )$, which renders \pref{pr:PFOne} to be non-convex. In what follows, we cast \pref{pr:PFTwo} to an equivalent problem, which can be optimized via a simple group-coordinate descent algorithm. We first state the following lemma that is given in \cite{Aflalo2011}.

\begin{lemma}
\label{lemm:EquivProblemNonConvexCase}

For $p \in [ \frac{1}{2}, 1 )$, $\boldsymbol{\lambda} \in \mathbb{R}^T$, $\boldsymbol{g} \in \mathbb{R}^T$, $\boldsymbol{g} \succeq \boldsymbol{0}$ and $\boldsymbol{g} \neq \boldsymbol{0}$, we have  

\begin{equation}
\min_{\boldsymbol{\lambda} \in \bar{B}_{\boldsymbol{\lambda}, q}} \boldsymbol{g}' \boldsymbol{\lambda}^{-1} = \nu(\boldsymbol{g})_p 
\label{pr:ANCOne}
\end{equation}

\noindent with $q \triangleq \frac{p}{1-p}$, and the optimal $\boldsymbol{\lambda}^*$ can be calculated as follows:

\begin{equation}
\boldsymbol{\lambda}^* = \left ( \frac{\boldsymbol{g}}{ \nu (\boldsymbol{g})_p } \right )^{1-p}
\label{eq:ANCTwo}
\end{equation}

\end{lemma}

\noindent Note that as $p \in [ \frac{1}{2}, 1 )$, the minimization problem in \pref{pr:ANCOne} is convex with respect to $\boldsymbol{\lambda}$. In the following lemma it is shown that, for $p \in (0, \frac{1}{2})$, a similar convex equivalency can be constructed.

\begin{lemma}
\label{lemm:EquivProblemNonConvexCase2}

For $p \in (0, \frac{1}{2})$, $\boldsymbol{\phi} \in \mathbb{R}^T$, $\boldsymbol{g} \in \mathbb{R}^T$, $\boldsymbol{g} \succeq \boldsymbol{0}$ and $\boldsymbol{g} \neq \boldsymbol{0}$, we have  

\begin{equation}
\min_{\boldsymbol{\phi} \in \bar{B}_{\boldsymbol{\phi}, 1}} \boldsymbol{g}' \boldsymbol{\phi}^{-\frac{1}{q}} = \nu(\boldsymbol{g})_p 
\label{pr:ANCOne2}
\end{equation}

\noindent with $q \triangleq \frac{p}{1-p}$, and the optimal $\boldsymbol{\phi}^*$ can be calculated as follows:

\begin{equation}
\boldsymbol{\phi}^* = \left ( \frac{\boldsymbol{g}}{ \nu (\boldsymbol{g})_p } \right )^{p}
\label{eq:ANCTwo2}
\end{equation}

\end{lemma}

\noindent
Note that \pref{pr:ANCOne2} is convex for $p \in (0, \frac{1}{2})$. The proof of \lemmaref{lemm:EquivProblemNonConvexCase2} is given in \sref{sec:EquivProblemNonConvexCaseProof} of the Appendix. The next lemma illustrates that, as $p\in (0, \frac{1}{2})$, even though \pref{pr:ANCOne} is not convex, it is equivalent to the convex optimization problem (\ref{pr:ANCOne2}).

\begin{lemma}
\label{lemm:EquivProblemNonConvexCase3}
Under the conditions of \lemmaref{lemm:EquivProblemNonConvexCase2}, we have that

\begin{equation}
\min_{\boldsymbol{\lambda} \in \bar{B}_{\boldsymbol{\lambda}, q}} \boldsymbol{g}' \boldsymbol{\lambda}^{-1}
\label{eq:equi_lambda}
\end{equation}

\noindent
is equivalent to \pref{pr:ANCOne2} with optimum solution

\begin{equation}
\boldsymbol{\lambda}^* = \left ( \frac{\boldsymbol{g}}{ \nu (\boldsymbol{g})_p } \right )^{1-p}
\label{eq:optimum_lambda}
\end{equation}
\end{lemma}

\noindent
The above lemma can be simply proved by letting $\boldsymbol{\phi} = \boldsymbol{\lambda}^p$. It states the fact that \eref{pr:ANCOne} and \eref{eq:ANCTwo} are true not only in the case for $p \in [\frac{1}{2}, 1)$, when \pref{pr:ANCOne} is convex, but also for $p \in (0, \frac{1}{2})$, when \pref{pr:ANCOne} is not convex. This fact directly leads to the following theorem:

\begin{theorem}
\label{thm:EquivProblemNonConvexCase}

Let $q \triangleq \frac{p}{1-p}$. For $p \in (0, 1)$, \ac{MT-MKL} optimization \pref{pr:PFOne} is equivalent to the following problem

\begin{equation}
\begin{aligned}
\min_{\boldsymbol{x} \in \Omega\left ( \boldsymbol{x} \right ), \boldsymbol{\lambda} \in \bar{B}_{\boldsymbol{\lambda},q}} \;\; \boldsymbol{g}'\left( \boldsymbol{x}\right)  \boldsymbol{\lambda}^{-1}
\label{pr:ANCThree}
\end{aligned}
\end{equation}

\end{theorem}

\noindent 
In order to optimize \pref{pr:PFOne}, we can equivalently optimize \pref{pr:ANCThree}, where a simple group-coordinate descent method can be applied. In each iteration, given $\boldsymbol{\theta}$ and $\boldsymbol{\lambda}$, we first optimize with respect to $\left \{ \boldsymbol{f}, \boldsymbol{\xi}, \boldsymbol{b} \right \}$. This problem involves $T$ independent \ac{SVM} problems, which can be solved using existing efficient \ac{SVM} solvers, such as LIBSVM \cite{CC01a}. Then, the optimization problem with respect to $\boldsymbol{\theta}$ becomes:

\begin{equation}
\min_{\boldsymbol{\boldsymbol{\theta}}\in \bar{B}_{\boldsymbol{\theta}, s}} \sum_{m=1}^M \frac{1}{\theta_m} \sum_{t=1}^T \frac{\left \| f_m^t \right \|_{H_m}^2}{\lambda_t}
\label{eq:non_convex_opt_theta}
\end{equation}

\noindent
Note that $\boldsymbol{\lambda} \succeq \boldsymbol{0}$ is always satisfied due to the constraint on $\boldsymbol{\lambda}$. Therefore, the aforementioned problem can be readily solved by applying \lemmaref{lemm:EquivProblemNonConvexCase}, which supplies a closed-form solution for $\boldsymbol{\theta}$. Finally, \eref{eq:ANCTwo} provides a closed-form expression to update $\boldsymbol{\lambda}$.

The previous group-coordinate descent algorithm is justified as follows. Assume $s > 1$, as $p \in [\frac{1}{2}, 1)$, \pref{pr:ANCThree} is convex with respect to each of the three blocks of variables, namely, $\left \{ \boldsymbol{f}, \boldsymbol{\xi}, \boldsymbol{b} \right \}$, $\boldsymbol{\theta}$ and $\boldsymbol{\lambda}$. The convergence of the algorithm is therefore guaranteed in this case, based on \cite[Prop. 2.7.1]{Bertsekas1999}. This is majorly because the \ac{SVM} solver provides a unique solution for $\left \{ \boldsymbol{f}, \boldsymbol{\xi}, \boldsymbol{b} \right \}$, and the solutions for $\boldsymbol{\theta}$ and $\boldsymbol{\lambda}$ are also uniquely obtained based on \lemmaref{lemm:EquivProblemNonConvexCase}. For a detailed proof, we refer the readers to the proof of Theorem 4 in \cite{Kloft2011}, which is similar in spirit. When $p \in (0, \frac{1}{2})$, optimizing with respect to  $\left \{ \boldsymbol{f}, \boldsymbol{\xi}, \boldsymbol{b} \right \}$ and $\boldsymbol{\theta}$ is the same as the case for $p \in [\frac{1}{2}, 1)$. The only difference is that, \pref{pr:ANCThree} is not convex with respect to $\boldsymbol{\lambda}$. However, due to \lemmaref{lemm:EquivProblemNonConvexCase3}, minimizing it with respect to $\boldsymbol{\lambda}$ is equivalent to the convex problem (\ref{pr:ANCOne2}). Hence, similar to the case when $p\in[\frac{1}{2}, 1)$, we iteratively solve three convex problems with respect to $\left \{ \boldsymbol{f}, \boldsymbol{\xi}, \boldsymbol{b} \right \}$, $\boldsymbol{\theta}$ and $\boldsymbol{\lambda}$ respectively. The algorithm will be able to converge, again, because of \cite[Prop. 2.7.1]{Bertsekas1999} for the same reason as previously stated.  The algorithm is summarized in \aref{alg:Algorithm2}.

\begin{algorithm}
\caption{Algorithm when $p \in (0, 1)$}
\label{alg:Algorithm2}
\begin{algorithmic}
\STATE Choose $M$ kernel functions. Calculate the kernel matrices $\boldsymbol{K}_m^t$ for the $T$ tasks and the $M$ kernels. Initialize $\boldsymbol{\theta}_0 \in \bar{B}_{\boldsymbol{\theta}, s}$, $\boldsymbol{\lambda}_0 \in \bar{B}_{\boldsymbol{\lambda}, q}$, $\epsilon$, and $k=0$. 
\WHILE{The change of objective function value is larger than $\epsilon$}
\STATE Given $\boldsymbol{\theta}_k$, $\boldsymbol{\lambda}_k$, solve $T$ \text{SVM} problems, get $\boldsymbol{f}_k$;
\STATE Given $\boldsymbol{f}_k$, $\boldsymbol{\lambda}_k$, calculate $\boldsymbol{\theta}_{k+1}$ based on \lemmaref{lemm:EquivProblemNonConvexCase};
\STATE Given $\boldsymbol{f}_k$, $\boldsymbol{\theta}_{k+1}$, calculate $\boldsymbol{\lambda}_{k+1}$ based on \eref{eq:ANCTwo};
\STATE $k \gets k+1$; 
\ENDWHILE

\end{algorithmic}
\end{algorithm}

It is worth emphasizing that our proposed algorithm for $p < 1$ has the same asymptotic computational complexity compared to the algorithm for $p = 1$, which was introduced at the end of \sref{sec:AlgorithmConvex}. In each iteration, the algorithm for $p = 1$ involves $T$ \ac{SVM} optimizations and a closed-form computation for $\boldsymbol{\theta}$, while the algorithm for $p < 1$ adds only one more closed-form computation for $\boldsymbol{\lambda}$. Hence, the computational complexity in each iteration is dominated by the one of the \ac{SVM} optimizer. On the other hand, the algorithm for $p \geq 1$ involves solving a quadratic programming problem (\ref{pr:ACTwentyTwo}) and a closed-form calculation for solving \pref{pr:ACTwentyThree}. Therefore, in each iteration, the computational complexity is dominated by the one of the \pref{pr:ACTwentyTwo} solver.

\section{Experiments}
\label{sec:Experiments}

In order to assess the merits of the new framework, we experimented with a few, selected data sets. In this section we present the obtained results and discuss the effects of varying $p$.

\subsection{Experimental Settings}
\label{sec:Settings}

Our experiments involve $6$ multi-class problems, namely, Wall-Following Robot Navigation (\textit{Robot}), Image Segmentation (\textit{Segment}), Landsat Satellite (\textit{Satellite}), Statlog Shuttle (\textit{Shuttle}), Vehicle Silhouettes (\textit{Vehicle}) and Steel Plates Faults (\textit{Steel}) data set. All data sets were retrieved from the UCI Machine Learning Repository \cite{Frank2010}.  For all data sets, an equal number of samples from each class was chosen for training. Note that the original \textit{Shuttle} data set has seven classes, four of which are poorly represented; for this data set we only chose data from the other three classes. The attributes of all data sets were normalized so that they lie in $\left [0,  1 \right ]$. Finally, we modeled each multi-class problem as an \ac{MTL} problem, whose recognition tasks considered every possible combination of a class versus another class (\ie\ $\binom{c}{2}$ one-vs-one tasks for $c$ classes). 

Besides the UCI benchmark problems, we also performed experiments on two widely used multi-task data set, namely the \textit{Letter} and \textit{Landmine} data sets, which are introduced in detail later. For all experiments, $11$ kernels were pre-specified: the Linear kernel, $2^{nd}$-order Polynomial kernel and Gaussian kernels with spread parameter values $\left \{ 2^{-7}, 2^{-5}, 2^{-3}, 2^{-1}, 2^{0}, 2^{1}, 2^{3}, 2^{5}, 2^{7} \right \}$. The value of \ac{SVM}'s parameter $C$ was selected via cross-validation. Also, the value of parameter $s$, appearing in the norm constraint of $\boldsymbol{\theta}$, was held fixed to $1.1$. Finally, in order to discuss the effect of $p$, we allowed it to take values in $\left \{ 0.01, 0.02, 0.05, 0.1, 0.2, 0.5, 1, 2, 5, 10, 20, 50, \infty \right \}$.

\subsection{Effect of p on Objective Function Values}
\label{sec:ObjectiveValues}

\begin{figure}[htpb]
	\begin{center}
		\centering
		\includegraphics[width=0.5\textwidth]{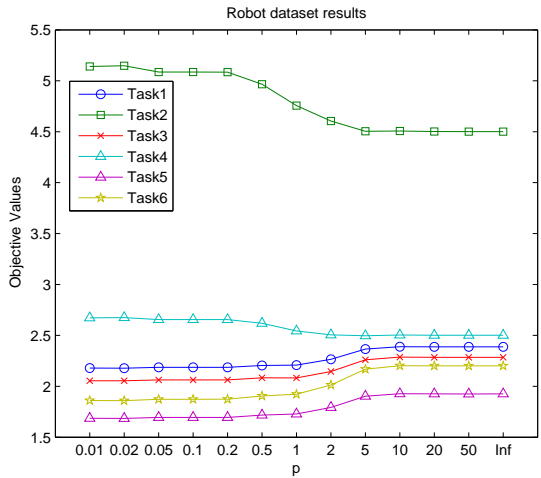}
		\caption{Experimental results for the \textit{Robot} data set. Objective function value of each task as a function of $p$.}
		\label{fig:objVal}
	\end{center}
\end{figure}

In \fref{fig:objVal}, we first show how the optimum objective function values of the $T$ tasks change as $p$ varies. In our subsequent discussion we'll focus on the results for the \textit{Robot} data set, due to its small number of tasks and, hence, is easy to analyze; similar observations can be made for the remaining data sets. 

\noindent{\bf Observations:} The optimum objective value of task $2$, which is the one exhibiting the highest objective value among the $6$ tasks, decreases, when $p$ increases, and achieves its lowest value for $p \geq 5$. Similar behavior is observed for task $4$, which is the task that has the second highest objective value. On the contrary, the other tasks have growing objective values as $p$ increases. Also, according to \propref{prop:ParetoPoints}, every $p$ value yields a \ac{PF} point of the relevant $T$-objective \ac{MOO} problem. This can be observed in the figure, since there is no $p$ such that the corresponding $T$ objective values lead to a dominating solution. 

\noindent {\bf Discussion:} The behavior displayed in \fref{fig:objVal} can be explained as follows. When $p \geq 1$, according to \lemmaref{lemm:EquivProblemConvexCase}, we know that $\boldsymbol{\lambda}^* \propto \boldsymbol{g}^{p-1}$. Therefore, the $\lambda_t$'s corresponding to tasks with high objective values are larger than the remaining $\lambda_t$'s. This means that these tasks are more heavily penalized as $p$ increases. In the extreme case, where $p \rightarrow \infty$, task $t_o$, that has the highest objective value, will have $\lambda_{t_0} = 1$ and the other $\lambda_t$'s will all be zero. This amounts to only task $t_0$ being penalized (thus optimized), while the other tasks' performances are ignored. Similarly, when $p < 1$, \lemmaref{lemm:EquivProblemNonConvexCase} and \lemmaref{lemm:EquivProblemNonConvexCase3} imply that $\boldsymbol{\lambda}^* \propto \boldsymbol{g}^{1-p}$. Therefore, in this case, the smaller $p$ is, the heavier the tasks with low objective values are penalized. This explains the trend of each curve in \fref{fig:objVal}. For the other data sets, similar observations can be stated for the same reasons.

\subsection{Effect of p on Classification Performances}
\label{sec:Classification}

\begin{figure}[htpb]
	\begin{center}
		\centering
		\includegraphics[width=0.5\textwidth]{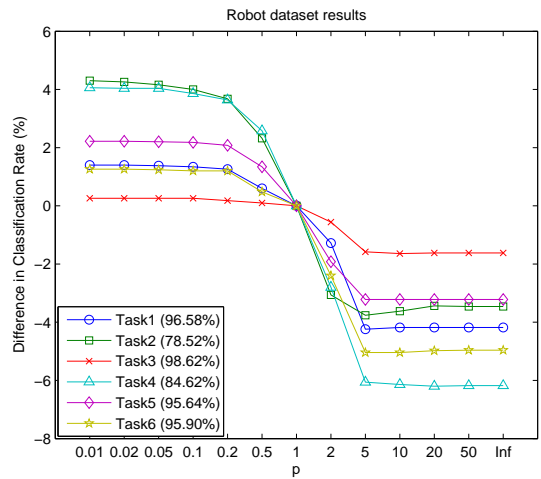}
		\caption{Experimental results for the \textit{Robot} data set. Difference in Classification Rate (DCR) for each task with changing $p$.}
		\label{fig:robotTaskwise}
	\end{center}
\end{figure}

\begin{figure}[htpb]
	\begin{center}
		\centering
		\includegraphics[width=0.5\textwidth]{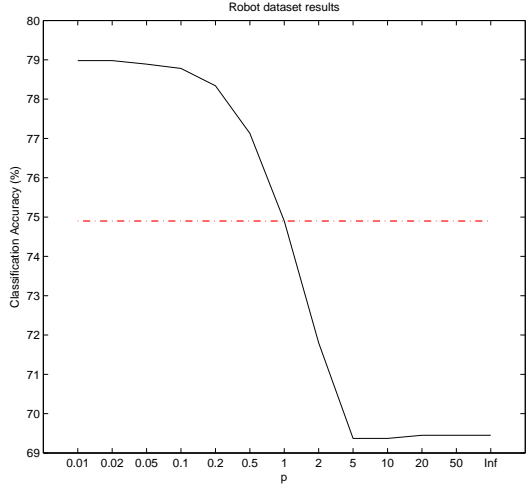}
		\caption{Experimental results for the \textit{Robot} data set. The black solid curve shows the overall multi-class classification accuracy as $p$ varies, while the red dashed line depicts the accuracy obtained, when $p=1$.}
  	\label{fig:robotOverall}
	\end{center}
\end{figure}

We now discuss how $p$'s value affects the classification performance of all tasks. Again, we analyze the performance on the \textit{Robot} data set in detail. \fref{fig:robotTaskwise} depicts the correct classification rate for each of the $6$ binary classification tasks, while \fref{fig:robotOverall} illustrates the overall classification rate as a multi-class problem. All experiments were performed using 5\% of the data available for training and the averages over $20$ runs are reported. Note that \fref{fig:robotTaskwise} showcases how the \textit{difference in classification rate} (DCR) changes with $p$. Here, DCR is defined as the difference in correct classification rates for each $p$ with respect to the rate when $p = 1$. The latter rate is displayed in parenthesis inside the legend for each task. As mentioned earlier, the latter case corresponds to the traditional \ac{MT-MKL} approach of optimizing the average of objective function values (all objectives are equally weighted). 

\noindent
{\bf Observations:} Upon inspection of \fref{fig:robotTaskwise}, one can immediately observe that the correct classification rate increases as $p$ decreases for each curve. Moreover, the best result is always achieved when $p < 1$ for all tasks. It can be seen that tasks $2$ and $4$, achieve the most significant improvement in classification rate, when $p<1$. On the other hand, the other four tasks also enjoy improved performance for $p < 1$, when compared to $p=1$. We used a t-test with significance level $\alpha = 0.05$ to compare the improvement between $p=1$ and $p=0.01$. The results of these tests confirmed the statistically significant improvement for all tasks except task $3$. \fref{fig:robotOverall} demonstrates the same behavior as the one shown in \fref{fig:robotTaskwise}. Note that the red dashed line shows the accuracy obtained when $p=1$. We immediately observe that the best performance is achieved as $p$ decreases towards $0.01$. Again, t-tests show statistically significant improvements. 

\noindent
{\bf Discussion:} The above observations can be explained as follows. The aim of the proposed \ac{MT-MKL} framework is to find an \ac{RKHS} by linearly combining pre-selected kernels via the coefficients $\boldsymbol{\theta}$, such that the $T$ tasks can achieve good performance. By applying \lemmaref{lemm:EquivProblemConvexCase} to \pref{pr:ACNine} and \pref{pr:ANCThree} for $\boldsymbol{\theta}$, one can easily see that the optimum solution is such that $\theta_m \propto (\sum_{t=1}^T \eta^t G_m^t )^{\frac{1}{s-1}}$, where we define $G_m^t \triangleq {\boldsymbol{\alpha}^t}' {\boldsymbol{Y}^t}' \boldsymbol{K}_m^t \boldsymbol{Y}^t \boldsymbol{\alpha}^t$ and $\eta^t \triangleq \lambda^t$, when $p \geq 1$, and $\eta^t \triangleq \frac{1}{\lambda^t}$, when $p < 1$. Also, as demonstrated earlier, we know that $\eta^t \propto (g^t )^{p-1}$ for $p > 0$. 

Based on these facts, let us first discuss the behavior as $p \rightarrow \infty$. In this scenario, as previously mentioned, the model is only optimizing task $t_0$, \ie\ the one with the highest objective value. It will hold that $\eta_{t_0} = 1$, while the other $\eta_t$'s will be $0$. In this case, the $\theta_m$'s are only determined by $G_m^{t_0}$. In other words, the \ac{RKHS} is learned only by the training samples from the $t_0$-th task, which is a (potentially, very) small proportion of the union of task-specific training sets. Hence, it is unsurprising that the correct classification rate is so low in this case. 

When $p$ is finite but very high, the $\eta_t$'s that correspond to tasks with high objective values will be much higher than the other $\eta_t$'s. As $p$ decreases, the former $\eta_t$'s will decrease in value. On the other hand, the other tasks will have increased $\eta_t$'s. This means that the $\theta_m$'s are estimated not only based on the training data for the tasks with high objective values, but also based on the training samples from other tasks. Therefore the classification accuracy will be improving for each task. This behavior continues as $p$ decreases towards $1$, until all $\eta_t$'s are equal. However, even though $\theta_m$ is now determined by the average of $G_m^t$'s, it does not necessarily mean that the training data of each task have the same influence on estimating the $\theta_m$'s. The reason is that the tasks with high objective values usually have much larger $G_m^t$ compared to the other tasks due to smaller \ac{SVM} margins (note that $G_m^t = \left \| f_m^t \right \|_{H_m}^2$). Thus, even though $\theta_m$ is determined by the average of $G_m^t$'s, the tasks with high objective values will influence this average value the most. One can interpret this as $\theta_m$ being estimated primarily based on the training data of these high objective valued tasks and to a lesser degree based on the training data of the other tasks.

Considering these last remarks, it is now straightforward to explain, why the classification performances obtained via the use of the proposed \ac{MT-MKL} framework is better for $p < 1$. In this case, the $\eta^t$'s have larger values for tasks with lower objective values. Therefore, the $G_m^t$'s associated to the these tasks perform a role of ever increasing importance as $p$ decreases, until some threshold value of $p$, after which the $\eta^t G_m^t$'s have similar values for all $t$. This can be interpreted as the $\theta_m$'s being estimated by considering the whole training set from all $T$ tasks, which eventually leads to improved performance for all tasks involved.

\begin{table}[htpb]
\begin{center}
\caption{Comparison of multi-class classification accuracy between $p < 1$, $p = 1$ and $p  \rightarrow \infty$} 
\label{Table:results}
\begin{tabular}{lcccccc}
\toprule 
10\% & Robot & Sat & Vec & Steel & Seg & Shuttle \\
\midrule
$p < 1$ & 90.11 & 85.48 & 61.47 & 73.35 & 90.22 & 98.77 \\
$p = 1$ & \underline{88.63} & \underline{83.79} & 60.61 & 73.33 & \underline{88.56} & \underline{97.78} \\
$p \rightarrow \infty$ & 87.01 & 82.12 & 46.75 & 71.49 & 84.98 & 72.32 \\
\# tasks & 6 & 15 & 6 & 3 & 21 & 3 \\
\# tasks $\uparrow$ & 6 & 12 & 5 & 2 & 21 & 1 \\
\midrule
20\% & Robot & Sat & Vec & Steel & Seg & Shuttle \\
\midrule
$p < 1$ & 92.70 & 87.78 & 66.97 & 76.33 & 92.86 & 99.20 \\
$p = 1$ & 92.36 & \underline{85.43} & 65.83 & 75.82 & \underline{90.72} & \underline{98.74} \\
$p \rightarrow \infty$ & 91.51 & 83.55 & 50.80 & 72.74 & 87.71 & 76.03 \\
\# tasks & 6 & 15 & 6 & 3 & 21 & 3 \\
\# tasks $\uparrow$ & 6 & 12 & 5 & 3 & 19 & 1 \\
\midrule
50\% & Robot & Sat & Vec & Steel & Seg & Shuttle \\
\midrule
$p < 1$ & 96.12 & 90.56 & 73.07 & 79.30 & 94.76 & 99.28 \\
$p = 1$ & 96.03 & \underline{87.53} & \underline{70.05} & \underline{76.78} & \underline{92.66} & \underline{98.71} \\
$p \rightarrow \infty$ & 95.64 & 85.60 & 52.70 & 72.93 & 90.36 & 80.71 \\
\# tasks & 6 & 15 & 6 & 3 & 21 & 3 \\
\# tasks $\uparrow$ & 6 & 14 & 6 & 3 & 21 & 2 \\
\bottomrule
\end{tabular}
\end{center}
\end{table}

In \tref{Table:results}, we show the overall classification performance results for all data sets with different size of training set. The percentage of the available samples that are used for training are shown in the first column for each set of experiments ($10\%$, $20\%$, and $50\%$ of the training set). The trends in overall accuracy are similar to the ones of task-wise performances, when $p$ varies. As was the case with the \textit{Robot} data set, the classification accuracy always decreases as $p$ increases. Here, we only show three results for each data set, namely, the highest accuracy obtained when $p < 1$, the accuracy when $p = 1$ and the accuracy when $p \rightarrow \infty$. Since it is important to compare the performance between the cases of $p < 1$ and $p = 1$ (the traditional \ac{MT-MKL} approach), a t-test was employed to test the statistical significance of the accuracy improvements between the cases of $p<1$ and $p=1$. In the table, underlined numbers indicate the results that are statistically significantly worse than the corresponding ones when $p<1$. It can be seen that the results are consistent with our motivation and that they experimentally validate our analysis: using $p < 1$ always achieves significantly higher accuracy compared to the traditional \ac{MTL} approach. Rows labeled as ``\# tasks'' indicate the total number of tasks for each data set and rows labeled as ``\# tasks  $\uparrow$'' show the number of tasks that experience an increase in classification rate as $p$ decreases. It can be seen that for all data sets, most tasks improve in performance with decreasing $p$, which means that the best performance for these tasks is obtained when $p$ is close to $0$.

\subsection{Comparison to Other Approaches}
\label{sec:Comparison}


We also compared our framework to a baseline approach and three popular \ac{MTL} models, namely \textit{Sparse MTL} \cite{Rakotomamonjy2011}, \textit{Tang's method} \cite{Tang2009}, and \textit{Minimax MTL} \cite{Mehta2012}. The baseline method trains each \ac{SVM} task individually using \ac{MKL}, \ie, for each task $t$, the task-specific kernel coefficients $\boldsymbol{\theta}^t \triangleq [\theta^t_1, \cdots, \theta^t_M]'$ are learned using the constraint $\|\boldsymbol{\theta}^t\|_q \leq 1, \ q \geq 1,  \forall t$. Next, \textit{Sparse MTL} \cite{Rakotomamonjy2011} is a popular \ac{MTL} model, which utilizes the $L_p - L_q$ mixed-norm $\sum_{m=1}^M (\sum_{t=1}^T \| f_m^t \|^q)^{p/q}$ as a regularizer. By letting $0 < p \leq 1$, this choice aims to encourage sparser kernel representation across tasks than the case when $p = 1$. Following the setting in \cite{Rakotomamonjy2011}, in our experiments we use the ranges $0 < p \leq 1$ and $1 \leq q \leq 2$ to identify the parameters $p$ and $q$ via cross-validation. 

Next, \textit{Tang's method} \cite{Tang2009} employs a different information sharing strategy. Unlike \textit{Sparse MTL}, which relates tasks via the use of the mixed-norm, \textit{Tang's method} utilizes a partially shared kernel function for the same purpose. Specifically, the kernel function for each task is defined as $k^t \triangleq \sum_{m=1}^M (\xi_m + \gamma_m^t)k_m$ with constraints $\sum_{m=1}^M (\xi_m + \gamma_m^t) = 1, \forall t$; $\xi_m \geq 0, \gamma_m^t \geq 0, \forall m, t$; $\sum_{m=1}^M\sum_{t=1}^T \gamma_m^t \leq \beta$. Obviously, by letting $\gamma_m^t = 0, \forall m, t$, the kernel function becomes the same for all tasks, which is the setting that we used in our method. Similar to several other \ac{MTL} formulations, \textit{Tang's method} minimizes the average loss of the $T$ tasks. 
In our experiments, the parameter $\beta$ is selected via cross-validation.

The recently proposed \textit{Minimax MTL} \cite{Mehta2012} aims to guarantee a minimum level of accuracy for each task. Its motivation differs to that of our method, which considers improving the \ac{MTL} performance by traversing the Pareto path of the associated \ac{MOO} problem. Therefore, although similar, the two models are differently formulated: \textit{Minimax MTL} minimizes the maximum (or more generally, the $L_p$-norm) of the $T$ loss functions, while the $L_p$-norm of our model is applied to the $T$ objectives (both regularizer and loss function). In our experiments, we followed the experimental settings of \cite{Mehta2012}: a linear model, \ie, $f^t(\boldsymbol{x}) \triangleq {\boldsymbol{w}^t}' \boldsymbol{x}$, in conjunction with the hinge loss were utilized. Also, we used the model parameter values $\alpha = 0.1T$ and $0.2T$ (please refer to \cite{Mehta2012} for model details), and we report here the best results obtained. Two different regularizers were considered. The first regularizer was $\| \boldsymbol{W} \|_{tr}$, where $\boldsymbol{W} \triangleq [\boldsymbol{w}^1, \cdots, \boldsymbol{w}^T]$ and $\| \boldsymbol{W} \|_{tr}$ denotes the trace norm, \ie, the sum of the singular values of $\boldsymbol{W}$, as proposed by Argyriou, Evgeniou and Pontil (AEP) \cite{Argyriou2008}. The second regularizer was $\| \boldsymbol{v}^0 \|^2 + \frac{\lambda_1}{T} \sum_{t=1}^T\| \boldsymbol{v}^t\|^2$, where it is assumed that $\boldsymbol{w}^t$ is given as $\boldsymbol{w}^t = \boldsymbol{v}^0 + \boldsymbol{v}^t, \forall t$ and the problem is optimized over $\boldsymbol{v}^0$ and the $\boldsymbol{v}^t$'s. This regularizer was proposed by Evgeniou and Pontil (EP) \cite{Evgeniou2004}. 

\begin{table}[htpb]
\begin{center}
\caption{Comparison of multi-class classification accuracy between our method and four other models} 
\label{Table:results_compare_four_other_methods}
\begin{tabular}{lcccccc}
\toprule 
10\% & Robot & Sat & Vec & Steel & Seg & Shuttle \\
\midrule
Pareto & 90.11 & 85.48 & 61.47 & 73.35 & 90.22 & 98.77 \\
Sparse  & \underline{87.59} & 85.43 & 60.58 & 75.21 & 90.01 & 98.66 \\
Tang  & \underline{88.61} & 85.27 & 60.50 & 73.03 & \underline{88.65} & 98.06\\
MM-AEP  & 92.78 & \underline{83.06} & 71.89 & \underline{65.65} & 89.27 & \underline{97.22} \\
MM-EP  & 93.96 & \underline{80.94} & 68.31 & \underline{69.45} & \underline{85.57} & \underline{97.43} \\
Baseline & \underline{87.93} & \underline{82.46} & 60.05 & 72.98 & \underline{83.48} & \underline{74.81} \\
\midrule
20\% & Robot & Sat & Vec & Steel & Seg & Shuttle \\
\midrule
Pareto & 92.70 & 87.78 & 66.97 & 76.33 & 92.86 & 99.20 \\
Sparse  & \underline{91.81} & 87.99 & 67.58 & 76.44 & 92.77 & 99.08 \\
Tang  & 92.13 & 87.40 & 66.67 & 75.07 & \underline{91.17} & \underline{98.83} \\
MM-AEP  & 95.10 & \underline{84.63} & 75.07 & \underline{66.83} & \underline{91.30} & \underline{97.56} \\
MM-EP  & 96.59 & \underline{82.08} & 73.46 & \underline{72.45} & \underline{89.89} & \underline{97.72} \\
Baseline & \underline{91.45} & \underline{84.15} & 65.61 & 75.17 & \underline{85.30} & \underline{78.94} \\
\midrule
50\% & Robot & Sat & Vec & Steel & Seg & Shuttle \\
\midrule
Pareto & 96.12 & 90.56 & 73.07 & 79.30 & 94.76 & 99.28 \\
Sparse  & 95.77 & 92.18 & 74.85 & 78.84 & 94.49 & 99.66 \\
Tang  & 95.68 & 90.52 & 74.04 & 78.64 & 94.59 & 99.43 \\
MM-AEP  & 97.71 & \underline{86.85} & 78.80 & \underline{67.54} & \underline{92.55} & \underline{97.98} \\
MM-EP  & 98.73 & \underline{85.49} & 75.57 & \underline{75.26} & \underline{93.15} & \underline{98.37} \\
Baseline & \underline{95.09} & \underline{86.32} & \underline{69.89} & \underline{76.66} & \underline{86.32} & \underline{82.59} \\
\bottomrule
\end{tabular}
\end{center}
\end{table}

The experimental results are reported in \tref{Table:results_compare_four_other_methods}, where we compare our method, labeled \textit{Pareto} on the first row of each section of the table, and the four aforementioned methods. Note that for our method, we reported the best result achieved when $p < 1$. For the \textit{Minimax MTL} method, we reported results, when both the AEP and EP regularizers are applied (labeled \textit{MM-AEP} and \textit{MM-EP} respectively). Underlined values represent results that are worse than the ones of our method in a statistically significant sense.

First, by comparing the \ac{MTL} methods to the baseline, we observe that, utilizing \ac{MTL} is indeed beneficial in improving the classification performance, with only very few exceptions. In addition, our method always outperforms the baseline model. These facts validate the effectiveness of the \ac{MTL} methodology. Secondly, we observe that, in most experiments, our method outperforms the other models and many of the observed improvements are statistically significant. In other words, in practice, there are indeed multi-task classification problems that can be benefit from our model, for which the $L_p$-norm, instead of the average, of the $T$ objectives is minimized.

In the sequel, we report results on two multi-task data sets mentioned earlier in the text. The first one, namely the \emph{Letter} data set\footnote{Available at: \url{http://multitask.cs.berkeley.edu/}}, is a collection of handwritten words compiled by Rob Kassel of the MIT Spoken Language Systems Group. The associated \ac{MTL} problem involves 8 tasks, each of which is a binary classification problem for handwritten letters. The 8 tasks are: `C' \vs\ `E', `G' \vs\ `Y', `M' \vs\ `N', `A' \vs\ `G', `I' \vs\ `J', `A' \vs\ `O', `F' \vs\ `T' and `H' \vs\ `N'. Each letter is represented by a $8 \times 16$ pixel image, which forms a $128$-dimensional feature vector. The goal for this problem is to correctly recognize the letters in each task. For the experiments, due to the large size of the data set ($45679$ data in total), we randomly chose $200$ samples for each task.

On the other hand, the \emph{Landmine} data set\footnote{Available at: \url{http://people.ee.duke.edu/~lcarin/LandmineData.zip}} consists of 29 binary classification tasks. Each datum is a $9$-dimensional feature vector extracted from radar images that capture a single region of landmine fields. The $9$ features include four moment-based features, three correlation-based features, one energy ratio feature and one spatial variance feature \cite{Xue2007}. Tasks $1-15$ correspond to regions that are relatively highly foliated, while the other $14$ tasks correspond to regions that are bare earth or desert. The tasks entail different amounts of data, varying from 30 to 96 samples. The goal is to detect landmines in specific regions.

\begin{table}[htpb]
\begin{center}
\caption{Comparison of multi-task classification accuracy between $p < 1$, $p = 1$ and $p  \rightarrow \infty$}
\label{tab:ResultMultiTask}
\begin{tabular}{l l l l l}
\toprule
Letter & 10\% & 20\% & 50\%  \\
\midrule
$p < 1$ & 83.95 & 87.51 & 90.61 \\
$p = 1$ & \underline{81.45} & \underline{86.42} & 90.01  \\
$p \rightarrow \infty$ & 77.67 & 85.60 & 89.80  \\
\# tasks & 8 & 8 & 8 \\
\# tasks $\uparrow$ & 8 & 8 & 7 \\
\midrule
Landmine & 20\% & 30\% & 50\%  \\
\midrule
$p < 1$ & 69.59 & 74.15 & 77.42 \\
$p = 1$ & \underline{67.24} & \underline{71.62} & 76.96  \\
$p \rightarrow \infty$ & 65.01 & 69.00 & 74.06  \\
\# tasks & 29 & 29 & 29 \\
\# tasks $\uparrow$ & 29 & 28 & 27 \\
\bottomrule
\end{tabular}
\end{center}
\end{table}

\begin{table}[htpb]
\begin{center}
\caption{Comparison of multi-task classification accuracy between our method and four other models}
\label{tab:ResultMultiTask_compare_four_other_methods}
\begin{tabular}{l l l l l}
\toprule
Letter & 10\% & 20\% & 50\%  \\
\midrule
Pareto & 83.95 & 87.51 & 90.61 \\
Sparse & 83.00 & 87.09 & 90.65 \\
Tang & \underline{80.86} & \underline{82.95} & \underline{84.87} \\
MM-AEP & 83.92 & 86.77 & \underline{88.94} \\
MM-EP & \underline{81.73} & \underline{84.14} & \underline{86.43} \\
Baseline & \underline{81.33} & 86.39 & \underline{89.80} \\
\midrule
Landmine & 20\% & 30\% & 50\%  \\
\midrule
Pareto & 69.59 & 74.15 & 77.42 \\
Sparse & \underline{58.89} & \underline{65.83} & \underline{75.82} \\
Tang & \underline{66.60} & \underline{70.89} & 76.08 \\
MM-AEP & \underline{59.27} & \underline{61.06} & \underline{67.34} \\
MM-EP & 69.69 & 73.62 & 76.40 \\
Baseline & \underline{66.64} & \underline{71.14} & 76.29 \\
\bottomrule
\end{tabular}
\end{center}
\end{table}

The experimental settings for these two problems were the same as the ones used in the previous experiments, except that we did not use 10\% of the \emph{Landmine} data set for training, due to the small size of the data set; instead, we used 20\%, 30\% and 50\% of the available training data. The averages of classification accuracy of each task are displayed in \tref{tab:ResultMultiTask}.

We can observe from \tref{tab:ResultMultiTask} and \tref{tab:ResultMultiTask_compare_four_other_methods} that the experimental results on the two benchmark multi-task problems are, again, consistent with our previous analysis. It is worth noting that the improved performance for $p < 1$ is achieved without incurring additional computational cost, since, as stated in \sref{sec:AlgorithmNonconvex}, the algorithm for $p<1$ has the same asymptotic computational complexity per iteration as the one for $p=1$. 



\acresetall

\section{Conclusions}
\label{sec:Conclusions}

To recapitulate, this paper introduced a new \ac{SVM}-based \ac{MT-MKL} framework that co-trains binary classification tasks. The task objectives are forced to share a common kernel, which is inferred from the data via \ac{MKL}. Unlike traditional \ac{MTL} approaches, which only consider optimizing the average of objective functions, it can be shown that our framework optimizes a specific variety of (implicitly-defined) conic combinations of these functions. These combinations are parameterized by a parameter $p > 0$ and specialize to the average of task objectives, when $p=1$.

Our motivation to construct such a framework stemmed from the observation that the solution obtained by optimizing the average of objective functions coincides with a particular point (solution) on the \ac{PF} (the set of all non-dominated solutions) of a \ac{MOO} problem, which considers the simultaneous vector-optimization of all objective functions involved in the \ac{MTL} setting. In the case of our framework, we showed that the conic combinations it considers trace out a path on the \ac{PF} by varying $p$. This path embodies the trade-off in perceived importance among task objectives rather than among task-related classification performances. Since the traditional \ac{MTL} approach is largely a heuristic, we argued that, by considering other conic combinations (thus, other points on the aforementioned path), we may actually be able to uniformly improve the classification performances of all tasks even more.  

In order to generate solutions for the framework's formulation, through a series of equivalence results we had to distinguish between two cases: $p \geq 1$ (convex case) and $p \in (0,1)$ (non-convex case). For each of these two cases we derived algorithms, which are straightforward to implement, as they can leverage from existing solvers and closed-form solutions, which we derived.

Towards assessing the framework's merits, we performed a series of experiments on $6$ benchmark multi-class recognition problems, which were modeled as multi-task problems suitable to our framework. Additionally, $2$ multi-task data sets were also considered. The obtained experimental results demonstrated clear advantages of considering only the range $0 < p <1$ for all data sets. When compared to the traditional \ac{MTL} approach ($p=1$), these advantages were reflected in the simultaneous classification improvements for each task individually, as well as in the improvement of overall accuracy for the mutli-class recognition problems we considered. We justified this phenomenon as follows: when $p$ decreases towards $0$, the objectives with lower value become more important, which implicitly increases the effective number of training samples for all tasks and allows each task-specific model to improve its classification accuracy.

Future works involve formulating Pareto-Path \ac{MT-MKL} models based on other kernel machines, such as Kernel Ridge Regression \cite{Saunders1998}, One-class SVM \cite{Scholkopf2001}, Support Vector Domain Description \cite{Tax1999}, etc. It will be of interest to analyze the behavior of these models, when the Pareto-Path \ac{MT-MKL} framework is adapted. Also, it would be of interest to derive generalization bounds for this new \ac{MT-MKL} framework, so that its performance merits are better explained.

\section*{Acknowledgments}
\addcontentsline{toc}{section}{Acknowledgments}

C. Li acknowledges partial support from \ac{NSF} grant No. 0806931 and No. 0963146. Moreover, M. Georgiopoulos acknowledges partial support from \ac{NSF} grants No. 0963146, No. 1200566, and No. 1161228. Finally, G. C. Anagnostopoulos acknowledges partial support from \ac{NSF} grants No. 0647018 and No. 1263011. Any opinions, findings, and conclusions or recommendations expressed in this material are those of the authors and do not necessarily reflect the views of the \ac{NSF}.
\bibliographystyle{plainurl}
\bibliography{IEEE-TNNLS2012paperB}

\appendix

\acresetall

\subsection{Proof of \propref{prop:ParetoPoints}}
\label{sec:ProofParetoPoints}

Assume that $\boldsymbol{x}^*$ is not a Pareto Front solution. Then there is a feasible point $\boldsymbol{x}^{**}$ satisfying $g_t\left ( \boldsymbol{x}^{**} \right ) \leq g_t\left ( \boldsymbol{x}^* \right ), \; \; t \in Z_T$, while at least for one $t$ it holds that $g_t\left ( \boldsymbol{x}^{**} \right ) < g_t\left ( \boldsymbol{x}^* \right )$. This means that $\nu \left ( \boldsymbol{g} \left ( \boldsymbol{x}^{**} \right ) \right )_p < \nu \left ( \boldsymbol{g} \left ( \boldsymbol{x}^* \right ) \right )_p$, which contradicts our assumption that $\boldsymbol{x}^*$ is the optimal of \pref{pr:PFThree}.

\subsection{Proof of \thmref{thm:EquivProblemConvexCase}}
\label{sec:ProofThmEquivProblemConvexCase}


Due to \lemmaref{lemm:EquivProblemConvexCase}, \pref{pr:PFOne} (or \ref{pr:PFTwo}) is equivalent to

\begin{equation}
\min_{\boldsymbol{x} \in \Omega\left (  \boldsymbol{x}\right )} \max_{\boldsymbol{\lambda} \in \bar{B}_{\boldsymbol{\lambda},q}} \;\; \boldsymbol{\lambda}'\boldsymbol{g}\left ( \boldsymbol{x} \right )
\label{pr:ACFive}
\end{equation}

\noindent 
when $p \geq 1$, where $q \triangleq \frac{p}{p-1}$. Here we remind the readers that $\boldsymbol{x}$ encompasses all the \ac{SVM}-related variables $\boldsymbol{f}, \boldsymbol{\theta}, \boldsymbol{\xi}, \boldsymbol{b}$. Since the objective function above is jointly convex with respect to $\boldsymbol{f}, \boldsymbol{\theta}, \boldsymbol{\xi}, \boldsymbol{b}$ and concave with respect to $\boldsymbol{\lambda}$, while the feasible set of the above problem is convex and compact, the order of minimization and maximization can be interchanged based on the Sion-Kakutani \emph{min-max} theorem \cite{Sion1958}, which yields:

\begin{equation}
\max_{\boldsymbol{\lambda} \in \bar{B}_{\boldsymbol{\lambda},q}} \min_{\boldsymbol{x} \in \Omega\left (  \boldsymbol{x}\right )}  \;\; \boldsymbol{\lambda}'\boldsymbol{g}\left ( \boldsymbol{x} \right )
\label{pr:ACSix}
\end{equation}

\noindent 
The inner minimization problem can be partially converted to the dual form of the \ac{SVM} problem:

\begin{equation}
\begin{aligned}
 \max_{\boldsymbol{\lambda}} \min_{\boldsymbol{\theta}} \max_{\boldsymbol{\alpha}} & \;\; \tilde{\Phi}(\boldsymbol{\theta}, \boldsymbol{\alpha}, \boldsymbol{\lambda}) \\
& \text{s.t.} \; \; {\boldsymbol{\alpha}^{t}}' \boldsymbol{y}^t=0, \; \; \boldsymbol{0}\preceq \boldsymbol{\alpha}^{t}\preceq C\boldsymbol{1}, \; \; t \in N_T; \\
&  \; \; \boldsymbol{\theta} \in \bar{B}_{\boldsymbol{\theta}, s}, s \geq 1; \; \; \boldsymbol{\lambda} \in \bar{B}_{\boldsymbol{\lambda}, q}
\end{aligned}
\label{pr:ACSeven}
\end{equation}

\noindent 
where $\tilde{\Phi}(\boldsymbol{\theta}, \boldsymbol{\alpha}, \boldsymbol{\lambda}) = \sum_{t=1}^T \lambda^t ( {\boldsymbol{\alpha}^{t}}' \boldsymbol{1} - \frac{1}{2}{\boldsymbol{\alpha}^{t}}'{\boldsymbol{Y}^{t}}' $ $ (  \sum_{m=1}^M  \theta_m \boldsymbol{K}_m^t ) \boldsymbol{Y}^t \boldsymbol{\alpha}^{t} )$,  $\boldsymbol{y}^t \in \left\{ -1, 1\right\}^{N_t}$ is the vector containing all labels of the training data for the $t$-th task, $\boldsymbol{Y}^t$ is a diagonal matrix with the elements of $\boldsymbol{y}^t$ on its diagonal, and $\boldsymbol{K}_m^t$ is the kernel matrix with elements $k_m\left ( \boldsymbol{x}_i^t,\boldsymbol{x}_j^t  \right )$. Since the function $\max_{\boldsymbol{\alpha}} \tilde{\Phi}(\boldsymbol{\theta}, \boldsymbol{\alpha}, \boldsymbol{\lambda}) $ and the feasible set of the above problem satisfy the Sion-Kakutani \emph{min-max} theorem \cite{Sion1958}, the order of the outer maximization and the minimization can be changed once again, which results in the following optimization problem:

\begin{equation}
\begin{aligned}
 \min_{\boldsymbol{\theta}} \max_{\boldsymbol{\lambda}, \boldsymbol{\alpha}} & \;\; \tilde{\Phi}(\boldsymbol{\theta}, \boldsymbol{\alpha}, \boldsymbol{\lambda}) \\
& \text{s.t.} \; \; {\boldsymbol{\alpha}^{t}}' \boldsymbol{y}^t=0, \; \; \boldsymbol{0}\preceq \boldsymbol{\alpha}^{t} \preceq C\boldsymbol{1}, \; \; t \in N_T; \\
&  \; \; \boldsymbol{\theta} \in \bar{B}_{\boldsymbol{\theta}, s}, s \geq 1; \; \; \boldsymbol{\lambda} \in \bar{B}_{\boldsymbol{\lambda}, q}
\end{aligned}
\label{pr:ACEight}
\end{equation}

\noindent 
Next, if we consider the change of variables $\beta_i^t \triangleq \alpha_i^t \lambda^t \;, t \in N_T$, then \pref{pr:ACEight} becomes \pref{pr:ACNine}. Note that, in fact, the constraint ${\boldsymbol{\beta}^{t}}'\boldsymbol{y}^t=0$ should have been $\frac{{\boldsymbol{\beta}^{t}}'\boldsymbol{y}^t}{\lambda^t}=0$ if $\lambda^t \neq 0$ and $\boldsymbol{\beta}^{t} = \boldsymbol{0}$ if $\lambda^t = 0$. However, these two constraints are equivalent, even in the case, when $\lambda^t = 0$, as we must have $\boldsymbol{\beta}^t = \boldsymbol{0}$, because of the constraint $\boldsymbol{\beta}^{t}\preceq\ \lambda^t \boldsymbol{1}$.

\subsection{Proof of \thmref{thm:SpecialMinProb1}}
\label{sec:ProofSpecialMinProb1}

The proof is separated into two parts. Initially, we prove the first part of the theorem, where $\Omega \left( \boldsymbol{\theta} \right) \triangleq \left \{ \boldsymbol{\theta} | \;  \boldsymbol{\theta} \in \bar{B}_{\boldsymbol{\theta}, s} \right \}$.

Notice that the minimization problem is convex, since $\Omega$ is a convex set and the objective function has a positive semi-definite Hessian matrix. Hence, it features a unique minimizer $\boldsymbol{\theta}^*$, which will be the unique stationary point (saddle point) of the problem's Lagrangian $\mathcal{L}$. Also, since $\left \{\boldsymbol{\theta} | \nu \left ( \boldsymbol{\theta} \right )_s \leq 1  \right \} = \left \{\boldsymbol{\theta} | \nu \left ( \boldsymbol{\theta} \right )_s^s \leq 1  \right \}$, we redefine $\Omega \triangleq \left \{ \boldsymbol{\theta} | \boldsymbol{\theta} \succeq \boldsymbol{0}, \; \nu \left ( \boldsymbol{\theta} \right )_s^s \leq 1 \right \}$. Notice that $\left ( \boldsymbol{\theta} \right )_s^s  = \left \| \boldsymbol{\theta} \right \|_s^s = \boldsymbol{1}' \boldsymbol{\theta}^s$. Additionally, \pref{pr:ACTwentyFour} is equivalent to

\begin{equation}
	\min_{\boldsymbol{\theta} \in \Omega} \left [ \boldsymbol{1}' \boldsymbol{\theta}^s - s \boldsymbol{\psi}' \boldsymbol{\theta} \right ] 
	\label{eq:APXprob1equiv}
\end{equation}

\noindent
after multiplying the original objective function by $s > 1$. If $\boldsymbol{\lambda} \succeq \boldsymbol{0}$ and $\mu \geq 0$ are the dual variables associated to the constraints $\boldsymbol{\theta} \succeq \boldsymbol{0}$ and $\left \| \boldsymbol{\theta} \right \|_s^s \leq 1$ respectively, then, for \pref{eq:APXprob1equiv}, $\mathcal{L}$ is given as

\begin{equation}
	\mathcal{L} \left ( \boldsymbol{\theta}, \boldsymbol{\lambda}, \mu \right ) = \boldsymbol{1}' \boldsymbol{\theta}^s - s \boldsymbol{\psi}' \boldsymbol{\theta} - \boldsymbol{\lambda}' \boldsymbol{\theta} + \mu \left ( \boldsymbol{1}' \boldsymbol{\theta}^s - 1 \right ) 
	\label{eq:APXprob1equivLagrangian}
\end{equation}

\noindent
Setting the Lagrangian's gradient with respect to $\boldsymbol{\theta}$ to $\boldsymbol{0}$ yields

\begin{equation}
	\boldsymbol{\theta} = \frac{1}{(\mu+1)^r}  \left ( \boldsymbol{\psi} + \frac{1}{s} \boldsymbol{\lambda} \right )^r
	\label{eq:APXprob1_eq1}
\end{equation}

\noindent
where $r \triangleq \frac{1}{s-1}$. Now, $\boldsymbol{\theta} \succeq \boldsymbol{0}$ implies via \eref{eq:APXprob1_eq1} that $\boldsymbol{\lambda} \succeq -s \boldsymbol{\psi}$. Since $\boldsymbol{\lambda} \succeq \boldsymbol{0}$ must also hold, we have that $\boldsymbol{\lambda} \succeq \max \left \{ \boldsymbol{0}, -s \boldsymbol{\psi} \right \} = \boldsymbol{0}$, as $\boldsymbol{\psi} \succeq \boldsymbol{0}$ by assumption. Given this range of $\boldsymbol{\lambda}$, $\mathcal{L}$ is maximized for $\boldsymbol{\lambda}^* = \boldsymbol{0}$. Thus, \eref{eq:APXprob1_eq1} becomes

\begin{equation}
	\boldsymbol{\theta} = \frac{1}{(\mu+1)^r} \boldsymbol{\psi}^r
	\label{eq:APXprob1_eq2}
\end{equation}

\noindent
In view of \eref{eq:APXprob1_eq2}, the last constraint, $\left \| \boldsymbol{\theta} \right \|_s^s \leq 1$, implies that $\mu \geq \left \| \boldsymbol{\psi} \right \|_{rs} - 1$. Given that $\mu \geq 0$ must also hold, we conclude that $\mu \geq \max \left \{ 0, \left \| \boldsymbol{\psi} \right \|_{rs} - 1) \right \}$. Finally, for this range of $\mu$, $\mathcal{L}$ is maximized for $\mu^* = \max \left \{ 0, \left \| \boldsymbol{\psi} \right \|_{rs} - 1) \right \}$ and, therefore, substituting $\mu^*$ into \eref{eq:APXprob1_eq2} finally yields the desired minimizer \eref{eq:ACTwentyFive}. Note that the results we found for $\boldsymbol{\theta}^*$, $\boldsymbol{\lambda}^*$ and $\mu^*$ satisfy the problem's \ac{KKT} conditions.

Next, we prove the second part of the theorem, where $\Omega \left( \boldsymbol{\theta} \right) \triangleq \left \{ \boldsymbol{\theta} | \boldsymbol{\theta} \succeq \boldsymbol{0}, \; \boldsymbol{\theta} \in \bar{B}_{\boldsymbol{\theta}, 1}  \right \}$. In this case, the Lagrangian is given as 

\begin{equation}
\mathcal{L}\left ( \boldsymbol{\theta},\boldsymbol{\lambda},\mu \right ) = \frac{1}{s} \boldsymbol{1}' \boldsymbol{\theta}^s -  \boldsymbol{\psi}' \boldsymbol{\theta} - \boldsymbol{\lambda}' \boldsymbol{\theta} + \mu\left ( \boldsymbol{1}'\boldsymbol{\theta}-1 \right )
\label{eq:APXprob1_eq3}
\end{equation}

\noindent By setting the gradient with respect to $\boldsymbol{\theta}$ to $\boldsymbol{0}$ yields

\begin{equation}
\boldsymbol{\theta} = \left ( \boldsymbol{\psi} + \boldsymbol{\lambda}-\mu\boldsymbol{1} \right )^{r}
\label{eq:APXprob1_eq4}
\end{equation}

\noindent Since $\boldsymbol{\theta} \succeq \boldsymbol{0}$, and the complementary slackness condition $\lambda_m \theta_m = 0$ must hold, it gives $\boldsymbol{\lambda} = \max\left \{ \boldsymbol{0},\mu\boldsymbol{1}-\boldsymbol{\psi} \right \}$. Plug it into \eref{eq:APXprob1_eq3}, we have 

\begin{equation}
\begin{aligned} 
\boldsymbol{\theta} &= \left ( \boldsymbol{\psi} + \max \left \{ \boldsymbol{0}, \mu \boldsymbol{1} - \boldsymbol{\psi} \right \} - \mu \boldsymbol{1} \right )^{r}  \\
&= \left ( \max \left \{ \boldsymbol{\psi}- \mu \boldsymbol{1}, \boldsymbol{0} \right \} \right )^{r} 
\end{aligned}
\label{eq:APXprob1_eq5}
\end{equation}

\noindent To maximize the Lagrangian with respect to $\mu$, the latter variable should be as small as possible. When combining the conditions $\mu \geq 0$ and $\left \| \boldsymbol{\theta} \right \|_1 \leq 1$, we reach the desired conclusion.

\subsection{Proof to \lemmaref{lemm:EquivProblemNonConvexCase2}}
\label{sec:EquivProblemNonConvexCaseProof}

First note that, when $p \in (0, \frac{1}{2})$, we have $\frac{1}{q} \in (1, +\infty)$. We first form the Lagrangian of \pref{pr:ANCOne2}:

\begin{equation}
\mathcal{L} (\boldsymbol{\phi}, \alpha) = \boldsymbol{g}' \boldsymbol{\phi}^{- \frac{1}{q}} + \alpha (\sum_{t=1}^T \phi^t - 1) - \sum_{t=1}^T \beta^t \phi^t
\label{eq:APXlemma3_eq1}
\end{equation}

\noindent
Setting the derivative of \eref{eq:APXlemma3_eq1} with respect to $\phi^t$ equal to $0$ yields:

\begin{equation}
\phi^t =  \left[ \frac{g^t}{(\alpha - \beta^t)q} \right]^p , \; \forall t
\label{eq:APXlemma3_eq2}
\end{equation}

\noindent
Obviously, as long as $g^t \neq 0$, we have $\phi^t \neq 0$ and therefore $\beta^t = 0$ due to the complementary condition $\beta^t \phi^t = 0, \; \forall t$. In this situation, we have that $\phi^t \propto (g^t)^p$. On the other hand, if $g^t = 0$, there must be $\phi^t = 0$. Therefore, in summary, we have $\boldsymbol{\phi} \propto \boldsymbol{g}^p$. Since we can observe that the optimum $\boldsymbol{\phi}^*$ must lie on the boundary, where $\sum_{t=1}^T \phi^t = 1$, we have 

\begin{equation}
\boldsymbol{\phi}^* = \frac{\boldsymbol{g}^p}{\sum_{t=1}^T (g^t)^p}
\label{eq:APXlemma3_eq3}
\end{equation}

\noindent
which completes the proof.

\end{document}